\pdfoutput=1
\documentclass[conference]{IEEEtran}
\usepackage{cite}
\usepackage{textcomp}
\usepackage{xcolor}
\def\BibTeX{{\rm B\kern-.05em{\sc i\kern-.025em b}\kern-.08em
		T\kern-.1667em\lower.7ex\hbox{E}\kern-.125emX}}
	
\usepackage{amsmath,amssymb,amsfonts}
\usepackage{graphicx}
\usepackage{caption}
\usepackage{subcaption}
\usepackage[colorlinks=true]{hyperref} 
\usepackage[nameinlink, noabbrev, capitalise]{cleveref} 

\graphicspath{ {./images/} }
\creflabelformat{equation}{#2#1#3}

\begin{document}

\IEEEoverridecommandlockouts
\IEEEpubid{\makebox[\columnwidth]{
		~\copyright{}2019 IEEE \hfill} \hspace{\columnsep}\makebox[\columnwidth]{ }}

\title{On Class Imbalance and Background Filtering in \\ Visual Relationship Detection}

\author{
	\IEEEauthorblockN{Alessio Sarullo}
	\IEEEauthorblockA{\textit{School of Computer Science} \\
		\textit{University of Manchester}\\
		Manchester, UK \\
		alessio.sarullo@manchester.ac.uk}
	\and
	\IEEEauthorblockN{Tingting Mu}
	\IEEEauthorblockA{\textit{School of Computer Science} \\
		\textit{University of Manchester}\\
		Manchester, UK \\
		tingting.mu@manchester.ac.uk}
}
	
\maketitle

\begin{abstract}
	In this paper we investigate the problems of class imbalance and irrelevant relationships in Visual Relationship Detection (VRD). State-of-the-art deep VRD models still struggle to predict uncommon classes, limiting their applicability. Moreover, many methods are incapable of properly filtering out background relationships while predicting relevant ones. Although these problems are very apparent, they have both been overlooked so far. We analyse why this is the case and propose modifications to both model and training to alleviate the aforementioned issues, as well as suggesting new measures to complement existing ones and give a more holistic picture of the efficacy of a model.
\end{abstract}

\section{Introduction}
In recent years there have been many advances in the field of Computer Vision. Deep end-to-end architectures \cite{krizhevsky_imagenet_2012, simonyan_very_2014, he_deep_2016} have proven time and time again to be very effective in extracting visual features from an image that can be used to classify visible objects \cite{girshick_fast_2015, ren_faster_2015, redmon_you_2016, liu_ssd:_2016} and even predict their shape \cite{he_mask_2017}. While object detection and segmentation is certainly an important step towards image understanding, it is not enough: even when performed flawlessly, it fails at capturing interactions between objects, which is a fundamental part in the process of scene understanding. This is why there has recently been an increasing focus on applying deep learning to what is commonly called Visual Relationship Detection (VRD) \cite{lu_visual_2016, xu_scene_2017, newell_pixels_2017, zellers_neural_2018}.

Performing both object and relationship detection on an image yields a set of $ \langle subject, predicate, object \rangle $\footnote{Note that the term ``object'' can have two meanings: an entity in an image or the third element of a relationship triplet. The appropriate meaning can always been inferred by context.} triplets, where $ subject $  and $ object $ are entities such as \textit{man} or \textit{car} and $ predicate $ describes the relationship that ties them together (such as \textit{behind} or \textit{riding}). These triplets can then be used to produce a graph where nodes represent objects and edges represent relationships. This is commonly referred to as a \textit{scene graph} and it provides a structured and explicit image representation. Scene graphs have been successfully used in tasks such as Image Retrieval \cite{johnson_image_2015}, Image Generation \cite{johnson_image_2018} and Visual Question Answering \cite{lu_r-vqa:_2018, teney_graph-structured_2017}. 

The most common way in literature to tackle VRD is reducing it to a multi-class classification problem, using deep neural networks for feature extraction. A standard architecture \cite{zhang_visual_2017, lu_visual_2016, li_vip-cnn:_2017, li_scene_2017, dai_detecting_2017, xu_scene_2017, yu_visual_2017, liang_deep_2017, jae_hwang_tensorize_2018, newell_pixels_2017, zellers_neural_2018} includes a convolutional neural network (CNN) to obtain a visual feature map and detect objects, followed by another module that uses the information provided by the visual one in order to predict predicates between object pairs, picking the most likely out of many possible predicate classes. However, the model's capability to predict the appropriate predicate can be undermined if the data distribution is skewed towards some classes (i.e., there is \textit{class imbalance}). This is an issue in many applications \cite{he_learning_2008} and VRD is no exception: failing to detect infrequent relationships means that only a subset of classes can actually be predicted (see \cref{tab:pred_miss}). This leads to many details about interactions between objects being lost, as the model will prefer the common and generic \textit{on} instead of a more fitting predicate such as \textit{riding} or \textit{walking on}. We believe the reason class imbalance has been overlooked in literature is that it does not seem to reflect in poor performance, which happens because the most commonly used evaluation metric (recall) is relatively insensitive to low accuracy on small classes.

We tackle class imbalance using cost-sensitive learning, which is a paradigm that has been shown to be effective in this kind of scenarios \cite{khan_cost-sensitive_2017, he_learning_2008}. This solution does not require significant alterations to the model, operating instead on the training loss. On top of ease of implementation and usage, our solution still allows the model to predict every relationship class present in the dataset on which the model is trained, in contrast to other methods that cluster classes together and can thus only predict more generic categories, hindering their expressive power \cite{yang_visual_2018}.
We suggest a complementary metric to the standard recall that is agnostic to the number of examples belonging to each predicate, thus being equally affected by every class.

\begin{table}
	\centering
	\begin{tabular}{c|c|c}
		IMP \cite{xu_scene_2017} & NM \cite{zellers_neural_2018} & P2G \cite{newell_pixels_2017} \\ 
		\hline
		64\% &            48\% &        36\% \\
	\end{tabular}
	\caption{Percentage of relationship classes that three state-of-the-art models fail to predict (i.e., with 0 recall) on VG-SGG \cite{xu_scene_2017}.}
	\label{tab:pred_miss}
\end{table}

\begin{figure*}[h]
	\centering
	\includegraphics[width=.25\textwidth]{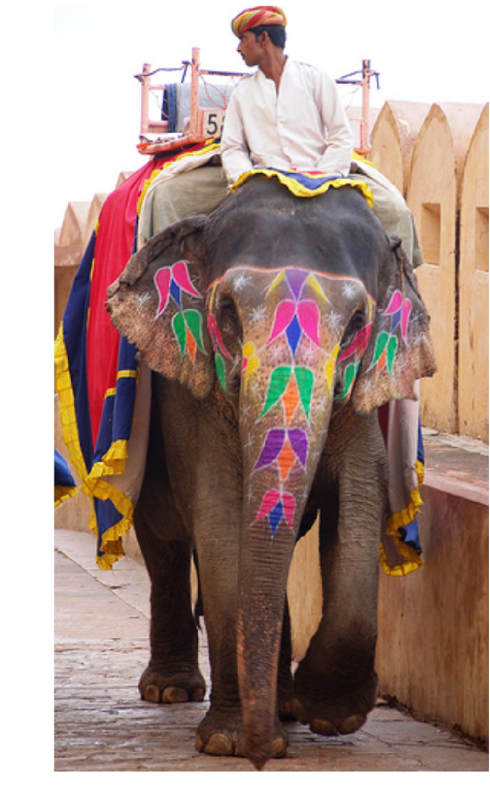}
	~~~~~~~~
	\includegraphics[width=.4\textwidth]{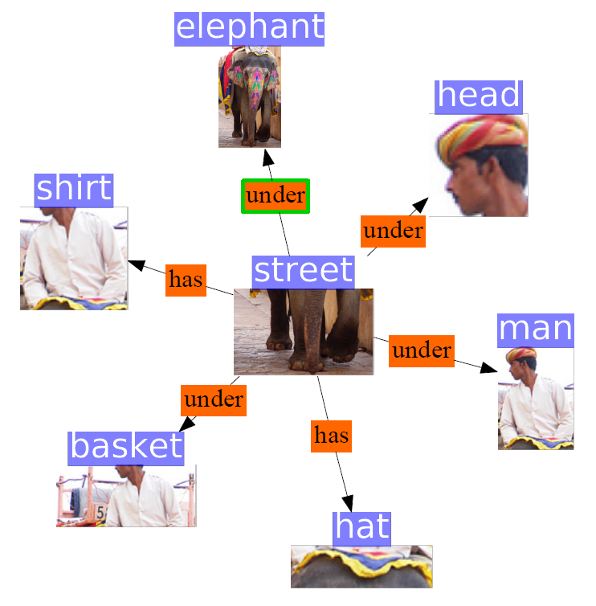}
	\caption{Every possible relationship whose subject is \textit{street}, as predicted by NeuralMotifs (or NM) \cite{zellers_neural_2018}. Objects have blue labels, relationship orange ones. The only meaningful predicate is highlighted in green, while the rest are irrelevant or absurd.}
	\label{fig:ex_bad}
\end{figure*}

We also analyse model's capability to predict \textit{null} relationships (also called \textit{background} or \textit{negative}), or equivalently abstaining from assigning any relationship for some subject-object pairs. Inability to do so will result in fully connected scene graphs, i.e., graphs where each object node is connected to every other. They represent an issue because the number of possible edges (relationships) grows quadratically in the number of nodes, but the majority of them are most likely just noise. This is shown in \cref{fig:ex_bad}: in the depicted image, a human annotator would probably not consider relevant any relationship that has \textit{street} as a subject except $ \langle street, under, elephant \rangle $,\footnote{A person would not describe the scene using $ street $ as a subject, but rather as an object. However, $ \langle street, under, elephant \rangle $ is still appropriate since it is the inverse of $ \langle elephant, walking~on, street \rangle $, which is the triplet corresponding to how a person would actually describe the image.} but the method proposed in \cite{zellers_neural_2018} predicts a predicate for every possible pair of objects. This example shows how the ability to predict the null relationship is important for a scene graph to constitute an appropriate description for the corresponding image. We propose an easy-to-implement architectural change that can be applied to any model to filter out noisy relationships and use the well-known $ F_1 $-score to measure its efficacy.

\section{Related work}

\subsection{Visual Relationship Detection} 
Recently, scene graphs have been successfully used in tasks such as Image Retrieval \cite{johnson_image_2015}, Image Generation \cite{johnson_image_2018} and Visual Question Answering \cite{lu_r-vqa:_2018, teney_graph-structured_2017}. The process of getting a scene graph from an image is called \textit{Scene Graph Generation} and VRD is a core part of it. Several deep learning methods have been proposed for this task \cite{zhang_visual_2017, lu_visual_2016, li_vip-cnn:_2017, li_scene_2017, dai_detecting_2017, xu_scene_2017, yu_visual_2017, liang_deep_2017, jae_hwang_tensorize_2018, newell_pixels_2017, zellers_neural_2018}. 

The most common architecture includes an object detector, usually Fast R-CNN \cite{girshick_fast_2015} or Faster R-CNN \cite{ren_faster_2015} with VGG \cite{simonyan_very_2014} or ResNet \cite{he_deep_2016} as a backbone. This detector is used to both localise objects in the image and extract their visual features. Sometimes the same CNN is used to extract features for \textit{visual phrases} \cite{li_vip-cnn:_2017}, i.e., parts of the image that contain a pair of possibly related objects. A few approaches try to predict both relationships and objects directly \cite{zhang_visual_2017, lu_visual_2016, yu_visual_2017}, while some other approaches use all of these visual features to build some kind of preliminary graph which is refined through message passing between nodes \cite{li_vip-cnn:_2017, li_scene_2017, xu_scene_2017, dai_detecting_2017}, with the goal of exchanging information across object and relationships. In \cite{liang_deep_2017, zellers_neural_2018} they incorporate context in a different manner, by computing a global embedding vector that is jointly used with region-specific features at prediction time. Differently from most methods, in \cite{newell_pixels_2017} they do not use any object detector. Instead, they compute a feature map and predict objects and relationships jointly for each mapped pixel, together with an embedding for each object. Objects are then assigned to each relationship based on embedding similarity with triplet components.

Some methods extend visual features with additional information. The most common way is to use pre-trained word embeddings \cite{mikolov_efficient_2013, pennington_glove:_2014}, which was first done in \cite{lu_visual_2016} to add a language prior that helps predicate detection by using the word embeddings of subject and object. Several methods have used the same idea since, including \cite{yu_visual_2017, zellers_neural_2018}. Some models \cite{zellers_neural_2018, jae_hwang_tensorize_2018} also exploit co-occurrence statistics found in the training set to obtain a prior predicate distribution over subject-object pairs that is then used to aid prediction by shifting probability away from unlikely pairs.

In our experiments we will use three of the most popular methods as baselines, namely \cite{xu_scene_2017}, \cite{newell_pixels_2017} and in particular \cite{zellers_neural_2018}, which represents the state of the art on VG-SGG at the time of writing.

\subsection{Class imbalance}
Class imbalance is an often overlooked issue in deep learning. A possible reason for this is that popular datasets like ImageNet \cite{russakovsky_imagenet_2015} or CIFAR \cite{krizhevsky_learning_2009} exhibit little to no class imbalance \cite{dong_class_2017}, thus making handling it unnecessary. However, recently there has been an increase in awareness of this issue in real-world datasets and a growing number of papers mention techniques to deal with this problem \cite{chung_cost-aware_2015, wang_training_2016, khan_cost-sensitive_2017, dong_class_2017, wang_learning_2017, huang_deep_2018}. All of these methods intervene on the training procedure as opposed to the data, since techniques such as oversampling or undersampling incur the risk of respectively promoting overfitting or removing useful information \cite{he_learning_2008}. Outside of this common denominator, the gamut of solutions is quite varied: in \cite{chung_cost-aware_2015} a cost-sensitive pre-training strategy is proposed that allows to extract features in a fair (i.e., balanced) way; in \cite{wang_training_2016} the authors use a cost-sensitive variation of the Mean Squared Error (MSE) loss function; \cite{khan_cost-sensitive_2017} propose to modify standard loss functions (MSE, hinge and cross-entropy) with a cost matrix that is learnt in alternation to the model on a separate validation set; in \cite{dong_class_2017} a triplet-based training procedure is used, where each batch is balanced as it contains both positive and hard negative examples; \cite{wang_learning_2017} propose a transfer learning approach to facilitate the prediction of small classes by using what the model has learnt on large ones; and in \cite{huang_deep_2018} the training is performed over quintuplets with the aim of clustering examples belonging to the same class together, so that infrequent classes can be distinguished from frequent ones.

However, to the best of our knowledge, only one very recent method \cite{yang_visual_2018} has explicitly tackled the problem of class imbalance in VRD. The authors apply K-means clustering on object labels based on their relationship frequency distributions on the training set, with the goal of reducing the number of labels and increase the amount of examples for each label. They apply their method to the \textit{Person In Context}\footnote{Available at \texttt{http://picdataset.com/challenge/index/}.} dataset, which focuses on human-centric relations (i.e., relationships where the subject is human) and accurate instance segmentation (as opposed to approximate bounding boxes). Our method differs in three main aspects: \textit{a)} it is applied to the relationship prediction module directly, instead of trying to improve it indirectly by clustering object classes; \textit{b)} it is easier to implementation and use, because it does not require significant changes to the model or training procedure; \textit{c)} it still allows the model to predict every relationship class present in the dataset, as opposed to only clusters of classes.

\section{Proposed methodology}

\subsection{Cost sensitive learning (CSL)} \label{sec:csl}
Our proposal to deal with class imbalance is cost-sensitive learning, which has proven effective in this kind of scenarios \cite{khan_cost-sensitive_2017, he_learning_2008}. Specifically, we propose the following cost-sensitive loss:
\begin{equation}
\label{eq:cs_bce_loss}
\begin{aligned}
L(\mathbf{Z}, \mathbf{T}) &\triangleq - \dfrac{1}{N} \sum_{i=1}^{N} \sum_{j=1}^{C} \ell(z_{ij}, t_{ij}) \\
\ell(z_{ij}, t_{ij}) &\triangleq u_{j} t_{ij} \log z_{ij} + v_{j} (1 -t_{ij}) \log(1 - z_{ij})
\end{aligned}
\end{equation}
where $ N $ is the number of examples, $ C $ is the number of predicate classes, $ \mathbf{Z}, \mathbf{T} \in \mathbb{R}^{N \times C} $ are respectively the matrices of predicted scores and 1-hot encoded targets, $ \boldsymbol{u}, \boldsymbol{v} \in \mathbb{R}^C $ are two vectors of class weights. The $ j $-th element of $ \boldsymbol{u} $ (i.e., $ u_j $) determines the importance given to positive examples \textit{of} class $ j $, while $ v_j $ weighs negative examples \textit{for} class $ j $. Thus, setting $ u_j > 1 $ tends to increase the recall for class $ j $, while $ v_j $ has an impact on the precision.

\begin{table}
	\centering
	\begin{tabular}{lr}
		&    eCE \\ \hline
		P2G with softmax \cite{newell_pixels_2017}          &  46.30 \\
		P2G with sigmoids                     				&  \textbf{21.66} \\
	\end{tabular}
	\caption{Comparison of eCE between the softmax model and the sigmoid one. Lower is better.}
	\label{tb:mce}
\end{table}

\begin{figure}
	\includegraphics[width=1\linewidth]{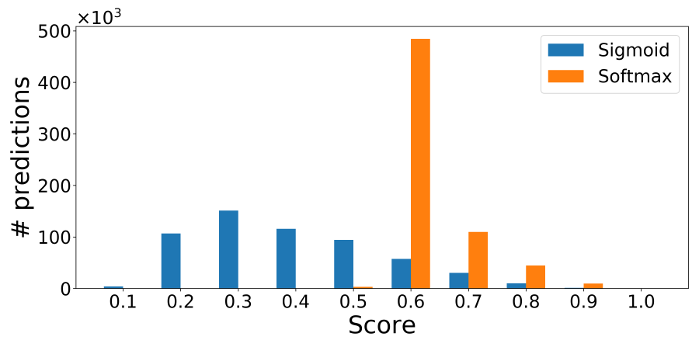}
	\caption{Comparison of quantized score distributions obtained with softmax and sigmoid activations using P2G \cite{newell_pixels_2017}.}
	\label{fig:p2g_score_hist}
\end{figure}

\begin{figure*}
	\centering
	\makebox[\textwidth][c]{\includegraphics[width=1.05\textwidth]{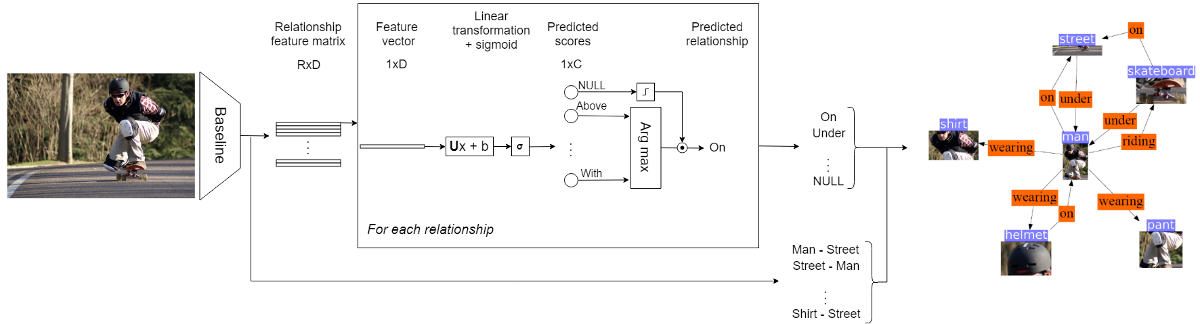}}
	\caption{Prediction pipeline with background filtering. In the picture $ R $ is the number of candidate relationships, $ D $ is the dimensionality of the feature space and $ C $ is the number of predicate classes.}
	\label{fig:fg_pred_pipeline}
\end{figure*}

The loss described in \cref{eq:cs_bce_loss} is a weighted version of the standard binary cross entropy loss, which is obtained when $ u_j = v_j = 1 \quad \forall j $. This loss derives from the maximum likelihood estimation of a sigmoid model, so we use sigmoid as activation function for the output layer. This comes with some additional benefits. First of all, predicates in VG-SGG semantically overlap with each other, as is the case, for instance, of \textit{on}, \textit{sitting-on}, \textit{on-back-of} or \textit{of}, \textit{belonging-to}. Using sigmoids has the advantage of modelling class predictions independently, which means that the output they produce preserves its meaning even with overlapping classes. Furthermore, it has been empirically shown \cite{hendrycks_baseline_2016} that softmax tends to produce inaccurate confidence scores. To verify that this applies in our case we compute the \textit{expected calibration error} (eCE) as defined in \cite{guo_calibration_2017}, which measures the difference between predicted score and expected accuracy. The eCE quantifies the intuitive idea that in a set of predictions with average score $ x $ we expect $100x\%$ of them to be correct. We report the results using the Pixels2Graph (P2G) model \cite{newell_pixels_2017} in \cref{tb:mce}. As it can be seen, sigmoid-produced scores are better calibrated than softmax ones by a large margin. This happens because the binary cross entropy loss used with sigmoids is more affected by negative samples, thus doing a better job at learning calibrated scores without the need of a larger amount of negative relationships to be provided. It can be indeed observed in \cref{fig:p2g_score_hist} that the histogram of scores computed using sigmoids is more widely distributed and shifted towards smaller values than the one obtained by a categorical cross entropy training.

We now describe how to set the two weight vectors $ \boldsymbol{u} $ and $ \boldsymbol{v} $ in the cost-sensitive binary cross entropy loss (\cref{eq:cs_bce_loss}). Following a framework first described in \cite{kukar_cost-sensitive_1998}, we first define the cost matrix $ \mathbb{R}^{C \times C} \ni \mathbf{W} \triangleq (w_{jk}) $, with
\begin{equation} \label{eq:cost_matrix}
w_{jk} \triangleq (1 - \delta_{jk}) \max\left(1, \log_2\dfrac{N_k}{N_j}\right )
\end{equation}
where $ \delta_{jk} $ is the Kronecker delta ($ \delta_{jk} = 1 $ if $ j = k $ and 0 otherwise) and $ N_j $ is the number of examples of class $ j $. Entry $ w_{jk} $ represent the costs of misclassifying an example of class $ j $ as belonging to class $ k $. We want this cost to be proportional to $ N_k $ because mistaking for a common class should be punished, as those classes are easier to predict due to the large number of examples. On the other hand, $ w_{jk} $ should also be inversely proportional to $ N_j $, because the few examples of an infrequent class $ j $ should not be mistaken for other classes. However, taking the unmodified ratio affects training negatively because of the very large gap between the two extremes (most common and least common classes), which is why we use the log ratio instead. Also, since the aim of this modification is to aid uncommon classes and not to hinder the learning of common ones, we do not allow the weights to be less than one, which would have a direct consequence on the training by reducing the gradient norm. Finally, the diagonal of the matrix is identically 0 because it represents the cost of correct classification, which should be null.

Given cost matrix $ \mathbf{W} $ we can derive the weight vectors we need to compute the loss. Let us assume we are considering example $ i $  of class $ c(i) $. Note that $ i $ contributes as a negative example for every class $ k \not= c(i) $, and its loss contribution should be proportional to the cost of misclassifying it for class $ k $, which is exactly $ w_{c(i),k} $. Therefore we set vector $ \boldsymbol{v} $ as
\begin{equation}
	\boldsymbol{v} = \mathbf{W}_{c(i), \cdot} \qquad \text{for an example of class } c(i)
\end{equation}
where $ \mathbf{W}_{k, \cdot} $ denotes the $k$-th row of matrix $ \mathbf{W} $. Coefficient $ u_{c(i)} $, on the other hand, defines the cost penalty for misclassifying example $ i $ so we set $ \boldsymbol{u} $ as the \textit{expected misclassification costs vector}:
\begin{equation} \label{eq:exp_miscl_cost}
u_j \triangleq \dfrac{1}{1 - P_j} \sum_{k \not= j} P_k w_{jk} \qquad \forall j = 1, \dots, C
\end{equation}
where $ P_j $ represents the prior probability of class $ j $, which we set as $ P_j \triangleq N_j / N $. Setting parameter vectors $ \boldsymbol{u} $ and $ \boldsymbol{v} $ this way increases the impact uncommon classes have on the training loss, thus making the classifier more likely to recognise them.

\subsection{Noisy relationships filtering (NRF)} \label{sec:bgpred}
Another issue that we tackle regards background relationships. The problem is again twofold: \textit{a)} some models are incapable of dealing with them and \textit{b)} the standard evaluation pipeline does not capture this issue.  

Models should be able to predict the null relationship. Not doing so leads to fully connected scene graphs, where each object node is predicted to have a relationship with every other. However, there is no guarantee that every object is related to all other ones: in fact, most of the possible edges in a scene graph do not convey any meaningful information, as shown in \cref{fig:ex_bad} for all the relationships having $ street $ as a subject. This problem could be solved by adding a module that distinguishes between relevant (foreground) and irrelevant (background) relationships on top of the standard classification architecture. We propose to implement this module by re-purposing the probability the network assigns to the background class to act as a filter for the output, as shown in \cref{fig:fg_pred_pipeline}. Using 0 as an index for the background class and $ \{1, \dots, C\} $ for the foreground ones, the decision rule is not $ \hat{y} \triangleq \arg\max_{j \in \{0, \dots, C\}} z_j $ any more, but rather
\begin{equation}
	\hat{y} \triangleq \mathbf{1}(z_0 < \theta) \arg\max_{j \in \{1, \dots, C\}} z_j \quad
\end{equation} 
where $ \theta $ is a threshold to be fixed (we use 0.5 in our experiments). Note that the background relationship prediction module is trained in a cost-sensitive way as well.

\begin{figure}
	\centering
	\includegraphics[width=1\linewidth]{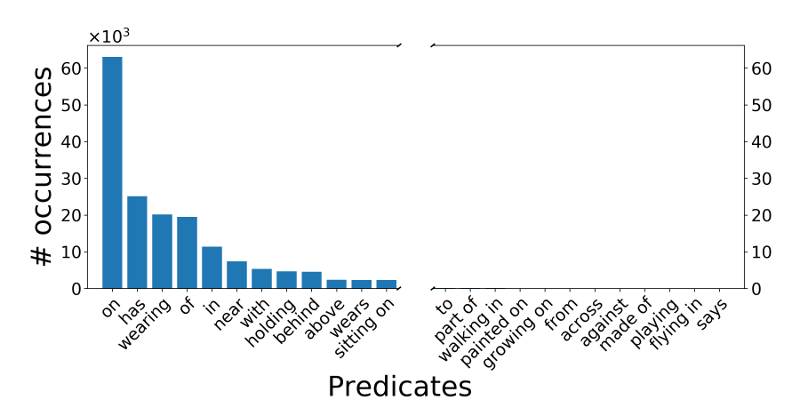}
	\caption{Number of examples per predicate in the VG-SGG test set \cite{xu_scene_2017} (the training set shows a very similar pattern). Predicates are sorted in a descending order. Only the 12 most common and least common ones are shown for readability reasons.}
	\label{fig:freq}
\end{figure}

\subsection{Evaluation metrics}
We believe that the main reason class imbalance has not been dealt with in VRD is that it does not appear to have an impact performance-wise. This happens because recall is the most commonly used evaluation metric and it is not affected much by low accuracy on small classes, as the total amount of examples it misses is very low compared to the correctly retrieved ones. We suggest the use of a metric called \textit{mean predicate classification recall} or \textit{mPCR}, defined as:
\begin{equation}
mPCR \triangleq \dfrac{1}{C}\sum_{j=1}^{C} \dfrac{1}{|\mathcal{G}(j)|} \sum_{\langle i, s, o \rangle \in \mathcal{G}(j)} \mathbf{1}\big(j = \hat{y}(i, s, o)\big)
\end{equation}
where $ \mathcal{G}(j) $ is the set of triplets $ \langle i, s, o \rangle $ such that $ j $ is the predicate class associated to the subject-object pair $ \langle s, o \rangle $ in image $ i $, $ \hat{y}(i, s, o) $ is the model prediction for such pair and $ \mathbf{1} $ is the indicator function that is 1 if its argument is true and 0 otherwise. The mPCR is essentially equivalent to the standard macro-averaged recall and it is insensitive to the number of examples of a particular class, which is why it is suited to measure the variety of predicates the model is actually able to predict.

Moreover, as we mentioned, most object pairs should not be related. In fact, in VG-SGG on average only about 6\% of the pairs is part of a relationship. While this percentage cannot be considered a gold standard due to the open-world assumption (i.e., the ground truth only contains a subset of the correct relationships), it is clear that the sparsity of relationship predictions is a desirable property of the model (see \cref{fig:ex_bad}). This is why we also use the well-known $ F_1 $ measure, since it depends both on the recall, which is widely used in the field of VRD, and on the precision, which penalises false positives. Models that cannot discriminate between relevant and irrelevant relationships will be penalised in their $ F_1 $ score.

\begin{figure}
	\centering
	\begin{subfigure}[c]{.4\textwidth}
		\centering
		\includegraphics[width=\textwidth]{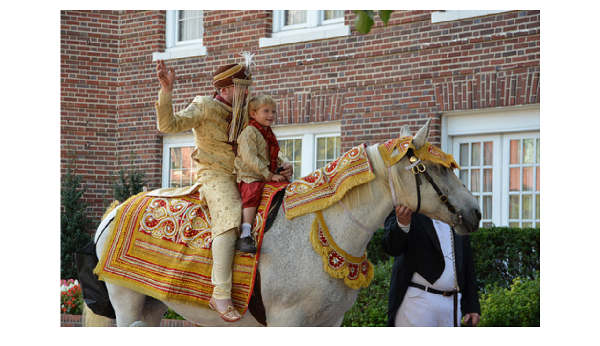}
		\caption{Image}
	\end{subfigure}
	
	\begin{subfigure}[b]{.2\textwidth}
		\centering
		\includegraphics[width=.9\textwidth]{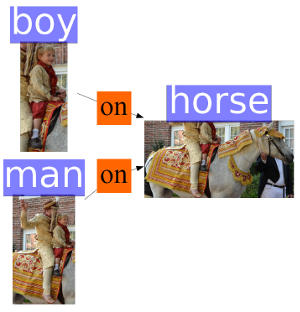}
		\caption{NM (baseline)}
	\end{subfigure}
	\begin{subfigure}[b]{.2\textwidth}
		\centering
		\includegraphics[width=.9\textwidth]{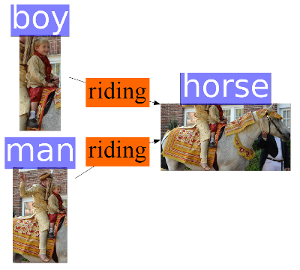}
		\caption{NM-CSL}
	\end{subfigure}		
	\caption{Example of relationships predicted by the baseline model (bottom left) and the same model trained through cost-sensitive learning (bottom right). A less common, but more appropriate predicate (\textit{riding}) is chosen by our model.}
	\label{fig:csl_pred_comparison}
\end{figure}

\section{Experiments and result analysis}

\begin{figure*}
	\centering
	\makebox[\textwidth][c]{\includegraphics[width=\textwidth]{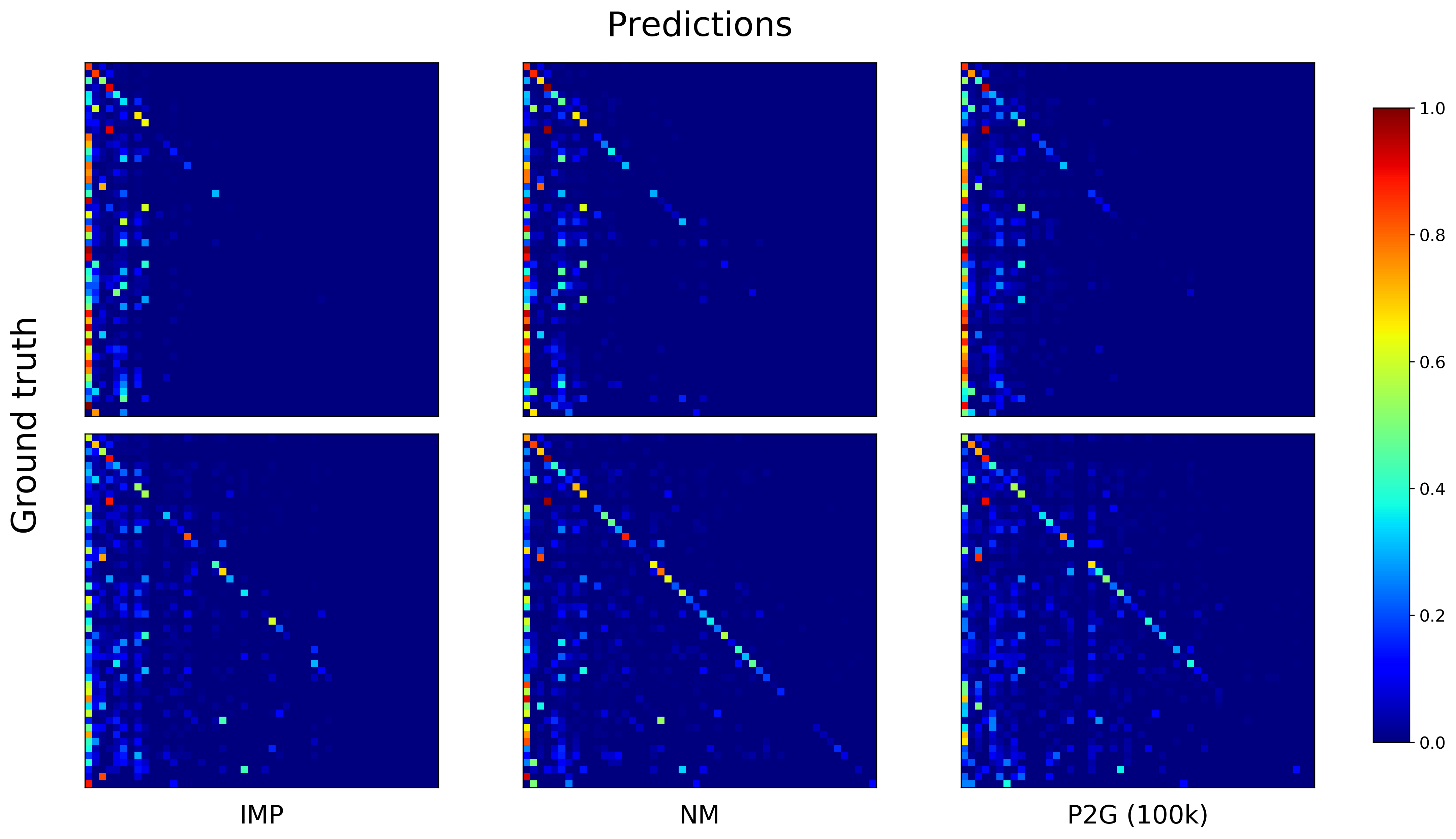}}
	\caption{Predicate confusion matrices obtained with the baseline model (top) and the modified one (bottom). Predicates are sorted from the most frequent to the least (as in \cref{fig:freq}) and the order is the same for both rows and columns and across methods.}
	\label{fig:csl_comparison}
\end{figure*}

\subsection{Experimental settings}
We applied our modifications on three models: Iterative Message Passing (IMP, \cite{xu_scene_2017}), Neural Motifs (NM, \cite{zellers_neural_2018}) and Pixels2Graph by Associative Embedding (P2G, \cite{newell_pixels_2017}). We tested them on VG-SGG \cite{xu_scene_2017}, the subset of Visual Genome \cite{krishna_visual_2017} most commonly used in the literature. It is a suitable dataset for cost-sensitive learning because its class distribution is very unbalanced (\cref{fig:freq}). All of our measures are computed considering the top 100 results produced by the model, to be aligned with the standard metric of recall@100 \cite{lu_visual_2016}.

\begin{table}
	\centering
	\begin{tabular}{lrrrr}
		&                 Recall &                   mPCR &                 Precision &                  $F_1$ \\
		\hline
		IMP \cite{xu_scene_2017}                 &                 52.11  &                  8.01  &                     5.11  &                  9.31  \\
		\hline
		IMP-CSL                                  &        \textbf{52.32}*  &        \textbf{16.97}  &                     5.27  &                  9.58  \\
		IMP-CSL-NRF                              &                 43.58  &                  9.94  &      \textbf{6.45} &        \textbf{11.22}  \\
		&&&& \\

		&                 Recall &                   mPCR &                 Precision &                  $F_1$ \\
		\hline
		NM \cite{zellers_neural_2018}            &        \textbf{67.02}  &                 15.06  &                     6.83  &                 12.40  \\
		\hline
		NM-CSL                                   &                 64.79  &        \textbf{29.03}  &                     6.55  &                 11.89  \\
		NM-CSL-NRF                               &                 60.03  &                 25.51  &           \textbf{13.12}  &        \textbf{21.52}  \\
		&&&& \\

		&                 Recall &                   mPCR &                 Precision &                  $F_1$ \\
		\hline
		P2G \cite{newell_pixels_2017} (2M iter)  &                 59.02  &                 14.42  &                    21.03  &                 31.01  \\
		\hline
		\hline
		P2G (100k iter)                          &        \textbf{52.84}  &                  9.41  &                    14.11  &                 22.27  \\
		\hline
		P2G-CSL (100k iter)                      &           49.42  &  \textbf{18.09}  &               7.58  &           13.14  \\
		P2G-CSL-NRF (100k iter)                  &                 45.28  &        15.05  &           \textbf{15.84}  &        \textbf{23.47}  \\
		
	\end{tabular}
	\caption{Summary of results. For each method the first sections (groups of rows) are dedicated to baselines, while the last one contains the proposed methodology. CSL indicates cost sensitive learning, NRF noisy relationship filtering. A star * denotes non-statistically significant improvements over the baseline.}
	\label{tb:all_res}
\end{table}

We generally use the settings provided by the authors to train the models. There are two main exceptions, both regarding P2G. 
The first is about training. The default schedule trains the network for 2 million iterations, which we estimate takes around three weeks on our machine. We reduce this to 100 thousands iterations, as at this point the model already produces sensible results. Since the focus of our experiments is assessing whether the modifications we propose are effective or not, our comparison is still fair, because we follow the same training regime for the model that includes our additions as for the corresponding baseline.
The second exception concerns P2G evaluation methodology. The authors allow multiple predictions for the same subject-object pair, but we disable this behaviour by only considering the best result for any single object pair at evaluation time. This way, the task remains the same (only one prediction per pair) across the different models we consider.

We have run each of our experiments 5 times and performed hypothesis tests to verify that our results are statistically significantly better than the baselines. In the comparison against IMP and NM we perform a one-sample Student's t-test, using as reference value the result obtained with the provided pre-trained models. For P2G we perform a Welch's t-test to take into account the possible difference in variance across experiments, since we train the baseline ourselves with a reduced schedule. All of our results are statistically significant at the 95\% confidence interval, unless otherwise specified.

\begin{table}
	\centering
	\begin{tabular}{lccc}
		& ~~~IMP~~~ 	& ~~~~NM~~~~ 	& P2G (100k) \\ 
		\hline
		Baseline 	& 64\% 	& 48\% 	& 52\% \\
		\hline
		CSL 		& 46\% 	& 14\% 	& 22\% \\
	\end{tabular}
	\caption{Percentage of predicate with 0 recall. Lower is better.}
	\label{tab:pred_miss_comparison}
\end{table}

\begin{figure*}
	\centering
	\makebox[\textwidth][c]{
		\subcaptionbox{Image}%
		[.2\textwidth]{\includegraphics[width=.2\textwidth]{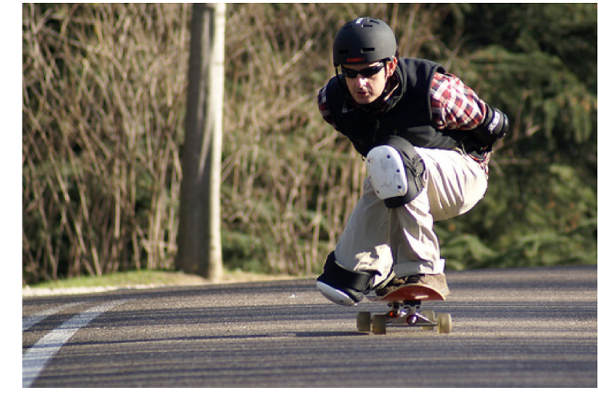}\vspace{1.5cm}}
		\subcaptionbox{Scene graph by NM \cite{zellers_neural_2018}}%
		[.4\textwidth]{\includegraphics[width=.4\textwidth]{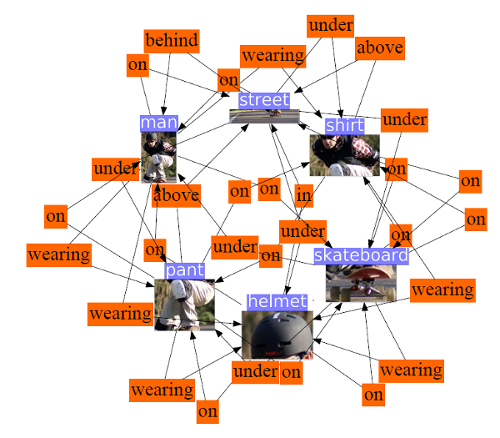}}
		\subcaptionbox{Scene graph by NM-CSL-NRF}%
		[.4\textwidth]{\includegraphics[width=.4\textwidth]{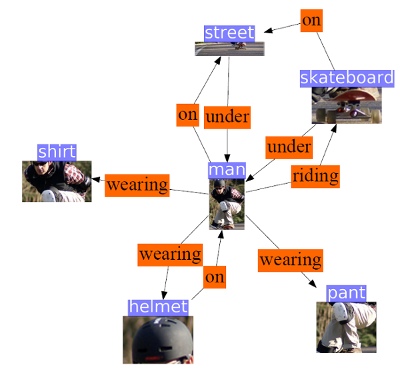}}	
	}
	\caption{Scene graph produced by the baseline model (middle) and our modified one (right). Noisy relationships are filtered out.}
	\label{fig:fg_pred_comparison}
\end{figure*}

\subsection{Analysis}
Our results are shown in \cref{tb:all_res}. As a general pattern, it can be observed that CSL makes the model significantly better at predicting uncommon classes and therefore reduces the effect of class imbalance during training, which is suggested by a higher mPCR. \Cref{tab:pred_miss_comparison} shows that the number of predicates never correctly predicted (i.e., they have 0 recall) decreases considerably. This can also be visually observed through confusion matrices (\cref{fig:csl_comparison}), that show more details about the distribution of predictions. However, CSL generally causes lower recall. This happens because the model now tries to predict infrequent classes in cases when it used to prefer common ones in non-cost-sensitive training, for example favouring \textit{riding} over a more generic \textit{on} when describing the relationship between a person and a horse (\cref{fig:csl_pred_comparison}).

When we also plug in background filtering recall and mPCR decrease, which is expected because fewer relationships are overall predicted as foreground. The advantage is that the model is now capable of discerning the difference between a relevant relationship and an irrelevant one, as an increased precision and $ F_1 $ score indicate. This can be seen in \cref{fig:fg_pred_comparison}: the baseline model outputs many relationships that are irrelevant or even incorrect, such as $ \langle street, under, pants \rangle $, $ \langle pants, on, shirt \rangle $ or $ \langle helmet, on, skateboard \rangle $. Our model, on the other hand, provides the relationships needed to describe the image and leaves out the noisy ones.

\section{Discussion and conclusion}
In this paper we examine with a critical eye current popular deep models for VRD such as the ones proposed in \cite{xu_scene_2017, newell_pixels_2017, zellers_neural_2018}. These are, or have been in the past, state-of-the-art models and exhibit advanced architectures that are able to effectively deal with a complex task as VRD. However, some commonly overlooked issues -- namely class imbalance and background filtering -- hamper their performance and limit their usefulness. Recall, which is the most commonly used evaluation metric, fails to detect such problems.

Our contribution is twofold: we provide some simple adaptations that can be easily plugged into any training procedure or model to alleviate class imbalance and background filtering issues and we also suggest the usage of some additional metrics to complement recall in performance evaluation.

A summary of our main results can be found in \cref{tb:all_res}. They suggest that
\begin{enumerate}
	\item cost-sensitive learning is effective at promoting uncommon class prediction, as shown by the increase in mPCR;
	\item background-filtering modules are effective at reducing the number of noisy relationships, thus increasing -- sometimes by quite a large margin -- the $F_1$ score.
\end{enumerate}

Although our approach gives promising result, it can be improved. The definition of cost matrix we use (see \cref{eq:cost_matrix}) captures the role that cost weights should fulfil, but it is somewhat heuristic. A few methods have been proposed in literature \cite{zhou_multi-class_2010, rudd_moon:_2016, khan_cost-sensitive_2017} to define cost-sensitive matrices. These can be integrated to further enhance performance and simplify design by not having to choose how to define the cost matrix. Another approach could be to complement our per-class cost-sensitive technique with per-example ones \cite{sarafianos_deep_2018}.

The background prediction module could also be enhanced. In training, a component could be added to the loss in order to put more emphasis on detective background examples from foreground ones, as proposed in \cite{shen_deepcontour:_2015}. From an architectural point of view, the module itself could be substituted by a more advanced one like that proposed in \cite{yang_graph_2018}, although at the cost of a more complex model.


\bibliography{library}{}
\bibliographystyle{ieeetr}

\end{document}